\documentclass[10pt,twocolumn,letterpaper]{article}

\usepackage{iccv}
\usepackage{times}
\usepackage{epsfig}
\usepackage{graphicx}
\usepackage{amsmath}
\usepackage{amssymb}
\usepackage{verbatim}
\usepackage{caption}
\usepackage{array}

\usepackage[pagebackref=true,breaklinks=true,letterpaper=true,colorlinks,bookmarks=false]{hyperref}

\iccvfinalcopy 

\begin{document}
\onecolumn
\begin{center}
{\LARGE IEEE Copyright Notice}
\begin {large}
\\~\\
\textsuperscript{\textcopyright}2019 IEEE. Personal use of this material is permitted. Permission from IEEE must be obtained for all
other uses, in any current or future media, including reprinting/republishing this material for advertising
or promotional purposes, creating new collective works, for resale or redistribution to servers or lists, or
reuse of any copyrighted component of this work in other works. 
\\~\\~\\
\textit{Accepted to be published in the 2019 OpenEDS Workshop: Eye Tracking for VR and AR at the International Conference on Computer Vision (ICCV), October 27- November 3, 2019, Seoul, Korea}
\end {large}
\end{center}
\twocolumn
\title{Assessment of Shift-Invariant CNN Gaze Mappings \\ for PS-OG Eye Movement Sensors}


\author{Henry K. Griffith, Dmytro Katrychuk, and Oleg V. Komogortsev\\
Texas State University, San Marcos, USA\\
{\tt\small \{h\_g169, d\_k139, ok\}@txstate.edu}
}

\maketitle

\begin{abstract}
Photosensor oculography (PS-OG) eye movement sensors offer desirable performance characteristics for integration within wireless head mounted devices (HMDs), including low power consumption and high sampling rates. To address the known performance degradation of these sensors due to HMD shifts, various machine learning techniques have been proposed for mapping sensor outputs to gaze location. This paper advances the understanding of a recently introduced convolutional neural network designed to provide shift invariant gaze mapping within a specified range of sensor translations. Performance is assessed for shift training examples which better reflect the distribution of values that would be generated through manual repositioning of the HMD during a dedicated collection of training data. The network is shown to exhibit comparable accuracy for this realistic shift distribution versus a previously considered rectangular grid, thereby enhancing the feasibility of in-field set-up. In addition, this work further demonstrates the practical viability of the proposed initialization process by demonstrating robust mapping performance versus training data scale. The ability to maintain reasonable accuracy for shifts extending beyond those introduced during training is also demonstrated.
 
\end{abstract}

\section{Introduction}

Eye movement (EM) sensors are a valuable tool for enhancing user experience in virtual and augmented reality (VR/AR) environments. This potential is evidenced through the recent acquisition activity of large technology companies, as well as the integration of EM sensors within newly released VR/AR platforms. By empowering gaze-based interaction, EM sensors can enhance user immersion in VR/AR ~\cite{piumsomboon_exploring_2017}. EM sensors also enable the implementation of gaze-contingent rendering, which reduces graphical resource requirements by exploiting the limitations of the human visual system ~\cite{padmanaban_optimizing_2017}.

EM sensors are traditionally implemented using video-based approaches. Denoted as video oculography (VOG), this technique estimates gaze location using computationally intensive image processing algorithms~\cite{hansen_eye_2009}. This process typically involves extracting the location of common eye features, such as the pupil and corneal glint, and then developing regression-based mappings to gaze location through a calibration procedure. Although VOG-based sensors are common in non-mobile applications, they are not well suited for deployment in wireless head-mounted devices (HMDs) due to their considerable computational overhead, which scales directly with the requisite sampling rate. This limitation is especially pertinent for emerging EM-enabled VR/AR applications, such as health assessment and eye-movement biometrics, which demand accurate estimates of gaze location at high sampling rates.

\begin{figure}[t]
\begin{center}
   \includegraphics[width=.9\linewidth]{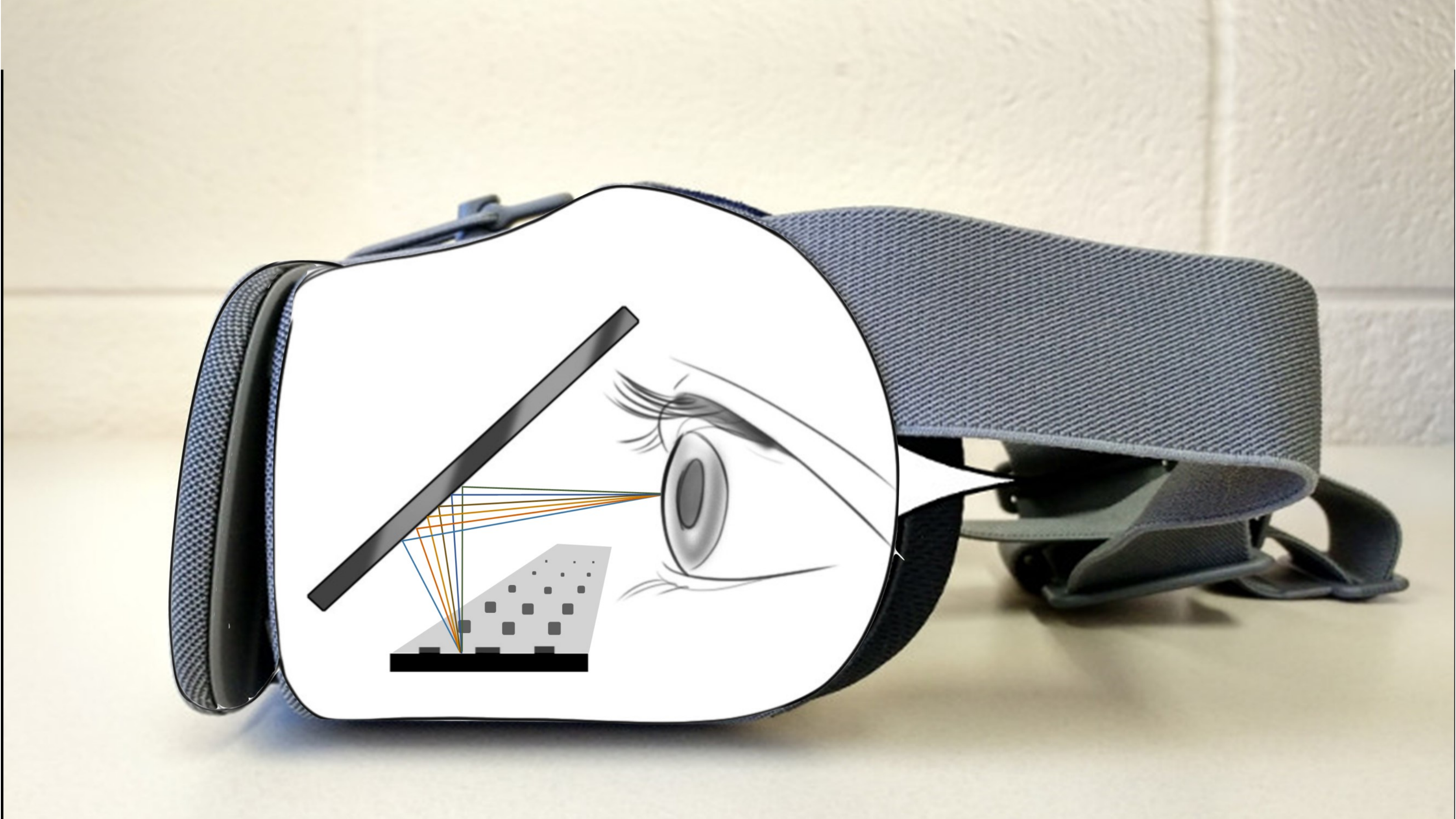}
\end{center}
\captionsetup{belowskip=-10pt}
\caption{Schematic Diagram of PS-OG Integration Using a Hot Mirror}
\label{fig:HMDInt}
\end{figure}

To meet the performance requirements of next-generation HMDs, alternative EM sensor architectures have been explored. Sensors employing photosensor oculography (PS-OG), which form gaze location estimates using variations in reflectivity captured by a discrete array of photosensors, offer considerable potential for meeting this demand. A diagram depicting the potential integration of a PS-OG sensor within an HMD is presented in Fig. \ref{fig:HMDInt}. As shown, a hot mirror is utilized in order to place the sensor array out of the user's line-of-sight.

PS-OG-based sensors suffer from considerable accuracy degradation in the presence of even slight sensor shifts, which induce relative displacements of the sensor's field of view with respect to the eye ~\cite{rigas_photosensor_2018}. Similar performance reductions have also been described for VOG-based sensors ~\cite{li_model-based_2008}. Shift-related performance degradation is especially concerning for untethered VR/AR applications, where increased user mobility will increase the likelihood of sensor movements.

\begin{figure*}[tb]
\begin{center}
   \includegraphics[width=\linewidth]{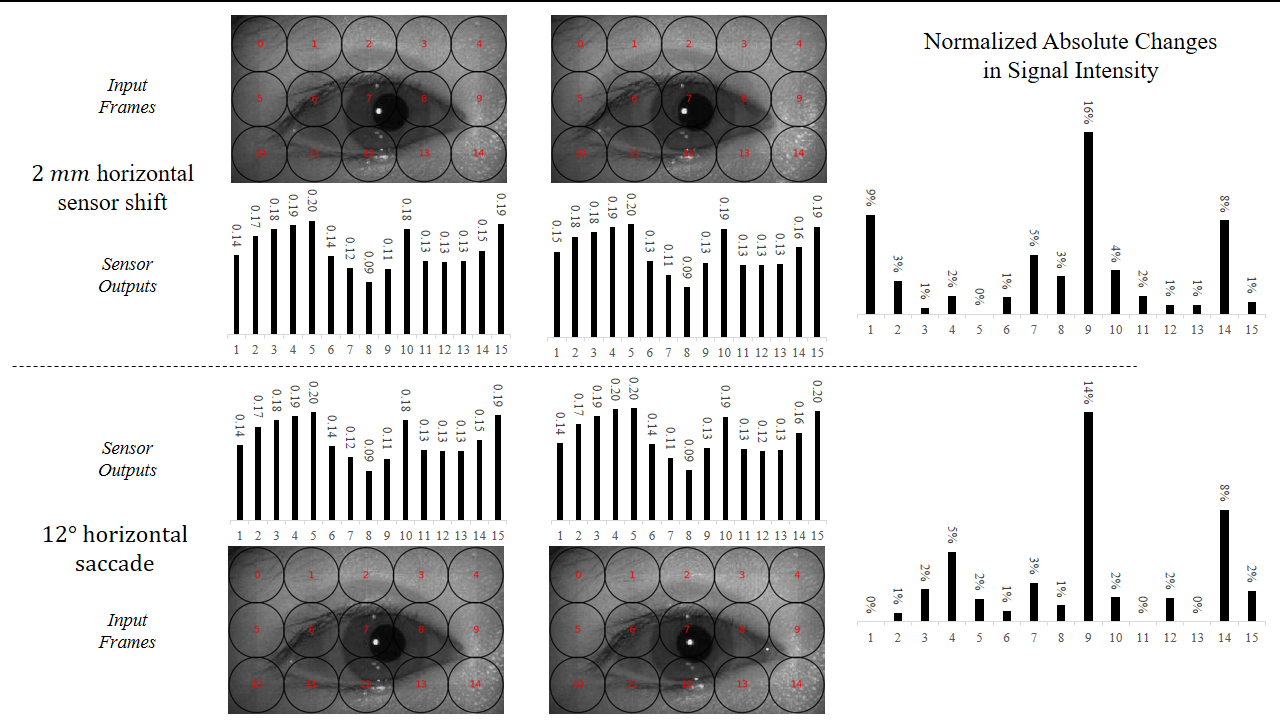}
\end{center}
   \caption{Variation in Signal Intensities Across the Array for Sensor Shifts and Gaze Displacements}
\label{fig:SensorOp}
\end{figure*}

The mechanism of shift-related performance reduction for PS-OG sensors is demonstrated in Fig. \ref{fig:SensorOp} using the simulation workflow utilized herein and described in Section 3.1. As the relative position of the array changes with respect to the eye, captured signal intensities vary considerably for a fixed gaze position. The resulting variation across the array is a function of the original field of view of each element, with those components located near strong reflection boundaries exhibiting the most significant changes. These perturbations are similar to those resulting from the rotation of the eye during transitions of gaze location. Absent of a correctional technique, the system may confuse these two scenarios, thereby degrading the spatial accuracy of gaze estimates. 

Machine learning models have been demonstrated to address this aforementioned scenario. To enable this approach, a mapping between raw sensor outputs and gaze location is learned over an anticipated range of sensor displacements. To generate the necessary training data,  prior work has suggested that the user manually reposition the HMD while fixating at varying target positions over the operating range of the sensor during a calibration procedure. These simulation-based studies have exclusively explored scenarios in which shift training data is available in an evenly spaced rectangular grid over the entire range of translations considered in testing. ~\cite{katrychuk_power-efficient_2019}, ~\cite{zemblys_making_2018}. 


While useful for establishing proof-of-concept, the considered spatial distribution of shift training data is inconsistent with what would be produced using the proposed generation mechanism. Moreover, although training data collection reduces user convenience by increasing initialization time, no analysis has been conducted to assess performance variability versus the amount of training data utilized. Such insight is valuable for practical deployment, as it allows developers to exploit potential trade-offs between accuracy and user convenience as a function of application requirements.

The research described herein seeks to address these limitations for a previously proposed convolutional neural network (CNN) gaze mapping ~\cite{katrychuk_power-efficient_2019}. The realism of the spatial distribution of available training data is enhanced by using Gaussian random variables to simulate manual repositioning of the HMD. Variability in accuracy as a function of available training data is explored, along with the capacity of learned mappings to generalize to shift ranges not observed during training. In addition, the range of shifts considered is expanded versus prior studies to the limits of the current simulation workflow. 

This manuscript extends the literature by demonstrating the ability of the gaze mapping network to achieve comparable accuracy to previous studies for a more realistic spatial distribution of shift training data. Namely, for an equivalent amount of training samples, best-case spatial accuracy was reduced by only 0.04$^{\circ}$ of the visual angle. Moreover, we show that achievable spatial accuracy is relatively robust to large variations in training data scale, with an 80\% reduction in the number of training samples yielding a 27\% degradation in spatial accuracy for subject-specific training. Finally, the ability of the network to generalize to out-of-sample shifts is shown. These observations are critical for supporting practical deployment of this approach in next-generation HMDs, where lengthy set-up procedures may be rejected by the user. 

\section{Related Work}
PS-OG sensors operate by actively illuminating the eye with infrared light, and then capturing subsequent reflections using an array of directive photosensors. Gaze location is estimated by exploiting variations in the captured reflectivity as the eye rotates. Reviewed work within this manuscript is limited to efforts attempting to enhance shift robustness of PS-OG sensors. A more thorough exploration of the underlying technology may be found in ~\cite{rigas_photosensor_2018}.

Rigas et al. demonstrated a hybrid technique in which a low speed (5 Hz) video sensor was used to correct errors in the PS-OG gaze estimate resulting from sensor shifts ~\cite{rigas_hybrid_2017}. Denoted as PS-V, this approach achieved a spatial accuracy of less than 1$^{\circ}$  of the visual angle for shifts in a $\pm$ 2 mm range. While PS-V may be suitable in certain scenarios, reliance on additional hardware is not ideal in resource-constrained environments such as wireless HMDs.

Zemblys and Komogortsev proposed a multilayer perceptron (MLP) network for learning a shift-invariant appearance-based mapping between sensor outputs and gaze location~\cite{zemblys_making_2018}. The proposed architecture was tested using a variation of the simulation framework originally introduced in ~\cite{swirski_rendering_2014}, using only the default parameters for the eye model and periocular region. A best-case spatial accuracy of 0.48$^{\circ}$ was achieved for shifts in a $\pm$ 1.75 mm range. Although this approach eliminates the video sensor requirement in practice, inference from the provided simulation results is limited by a lack of realism and diversity in the simulated eye model.  

To leverage the spatial structure of the PS-OG array output, Katrychuk et al. demonstrated a CNN for shift-invariant gaze mapping. In addition, the authors introduced a customized training approach, in which network weights were initialized through out-of-subject training, and then fine-tuned using subject-specific data. The network was tested on simulated senor outputs computed from real images obtained from a VOG system for multiple subjects. An accuracy of 1.07$^{\circ}$ was achieved for shifts in a $\pm$ 2 mm range for a CNN satisfying resource constraints intended to mimic those encountered in wireless HMDs~\cite{katrychuk_power-efficient_2019}.

While the proposed techniques in ~\cite{katrychuk_power-efficient_2019} were demonstrated to improve performance versus ~\cite{zemblys_making_2018} for an identical testing data set, both approaches utilized  a spatial distribution of shift training data which may not be realizable in a practical scenario. Namely, if requisite training data is generated through manual repositioning of the HMD by an end-user, the regularity, scale, and spatial distribution of available training data will vary substantially from the assumptions employed in these manuscripts, which considered only regularly-spaced rectangular distributions.

\section{Methods}
\subsection{Overview of Simulation Workflow}
All simulations conducted herein utilized a data set and simulation workflow nearly identical to that presented in \cite{rigas_hybrid_2017}. While summarized in this section, full process details are available in the original manuscript, with the entire modified code base and data set available online\footnote{\url{https://digital.library.txstate.edu/handle/10877/8574}}. 

Twenty-three subjects were recorded performing a random saccade task using a customized VOG eye tracker described in ~\cite{abdulin_study_2017}. All subjects provided informed consent under an experimental protocol approved by the Institutional Research Board at Texas State University. Before viewing the stimulus, participants performed a standard calibration procedure to ensure the validity of the recorded gaze estimates. 

\begin{figure*}[t]
\begin{center}
   \includegraphics[scale=0.4]{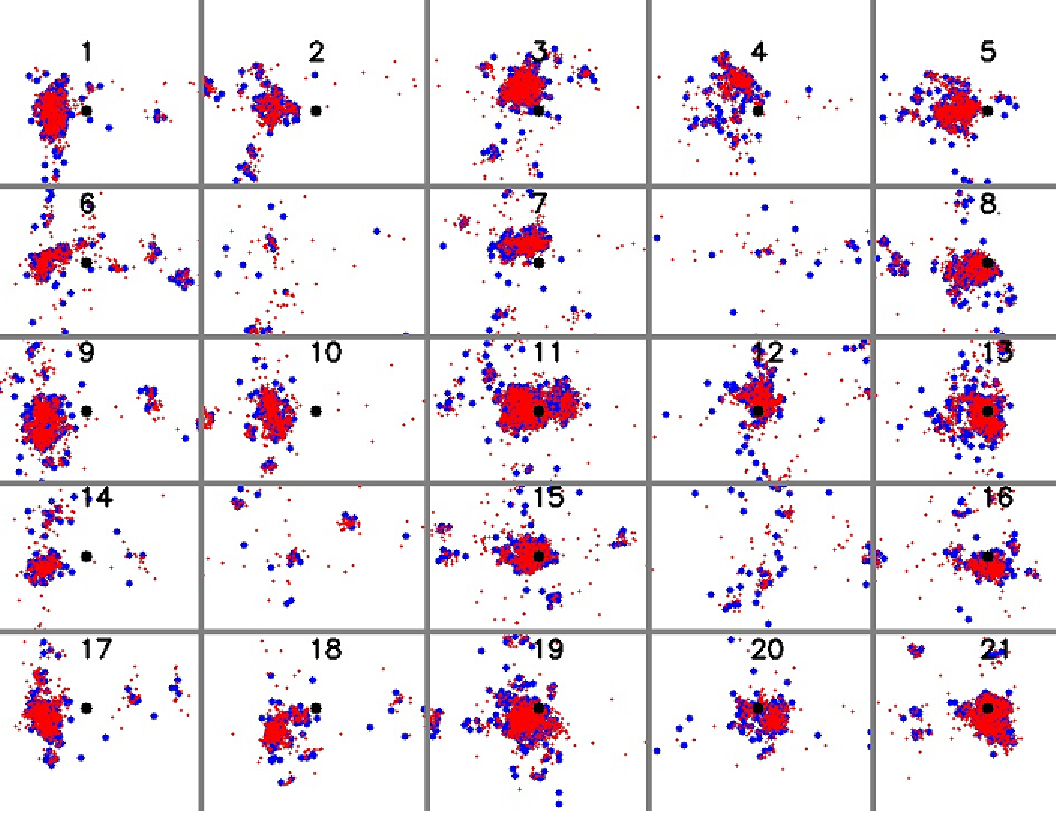}
\end{center}
\caption{Spatial Distribution of Stimulus Fixations (Black Dots) and            Corresponding Gaze Locations Across Screen (Blue and Red Dots Used to Denote Train+Validation and Test Split)}
\label{fig:DataDist}
\end{figure*}

The stimulus provided 25 fixations distributed across the operating range of the device ($\pm$ 20.51$^{\circ}$ and ± 16.7$^{\circ}$ of the visual angle in the horizontal and vertical directions, respectively). Each fixation was maintained for a random interval of time, yielding an asymmetric spatial distribution of data. Recorded VOG-based gaze estimates are utilized as the ground-truth values for all accuracy computations reported in the remainder of this manuscript. A visualization of these gaze locations for a specific experimental trial is provided  Fig. \ref{fig:DataDist}, where blue and red dots are used to denote the result of the random train/validation/test partition described in the forthcoming Machine Learning subsection. Fixation values induced by the stimulus are denoted with black dots and labeled with numerical identifiers. Gridlines are used to bin values within the contained areas for purposes of reporting spatial accuracy as specified in Section 4.1.  

A marker was placed on the nasal bridge of each participant in order to compute head movements during the recording process. Planar sensor translations were simulated using a customized cropping strategy which accounted for these movements. Images were cropped by a fixed pixel amount corresponding to the desired shift after adjusting for the determined head movement. The relationship between image pixels and length was determined by imaging a ruler at the beginning of the recording process. 

\begin{figure}
\begin{center}
   \includegraphics[width=.9\linewidth]{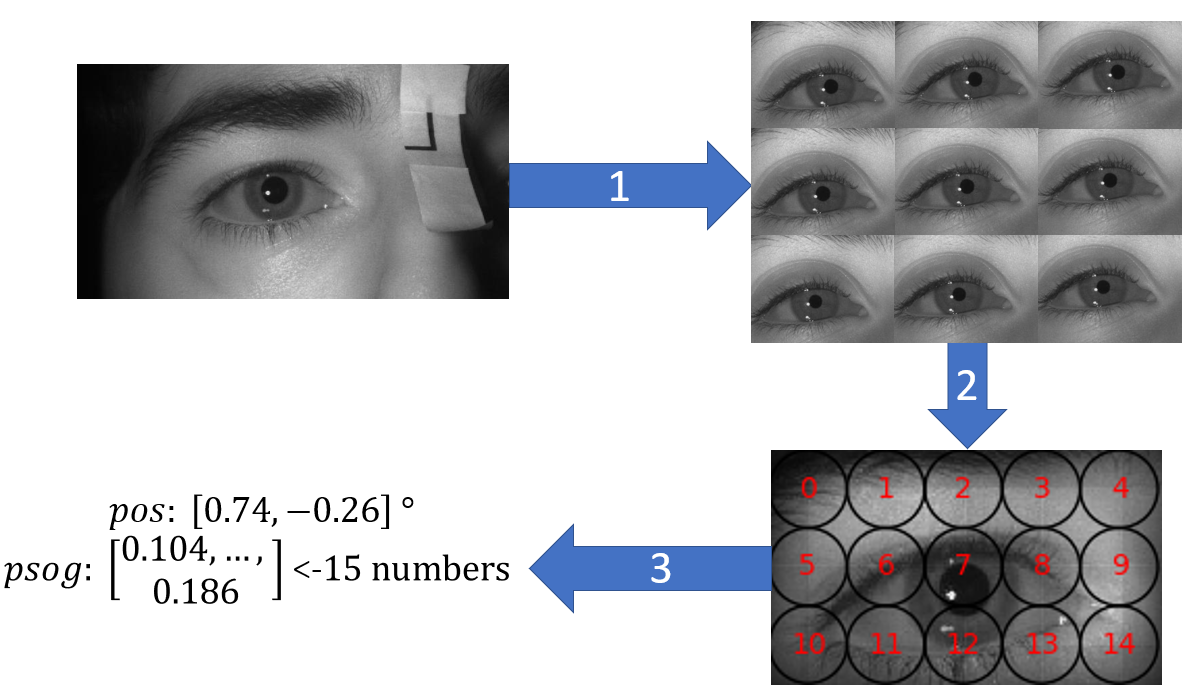}
\end{center}
\captionsetup{belowskip=-10pt}
\caption{Simulation Workflow}
\label{fig:sim_workflow}
\end{figure}

A 3 x 5 rectangular array of photosensors was simulated using a procedure originally introduced in~\cite{zemblys_making_2018}. The receptive field of each device was modeled using a Gaussian distribution (zero mean, standard deviation equivalent to $\frac{1}{4}$ of the window size). The output of each sensor was determined by convolving this distribution with a fixed 121 pixel window of the raw image. The simulation workflow is summarized in Fig. \ref{fig:sim_workflow}.

\subsection{Shift Generation Mechanism}
\begin{figure}[t]
\begin{center}
   \hspace*{1cm}\includegraphics[scale=0.5]{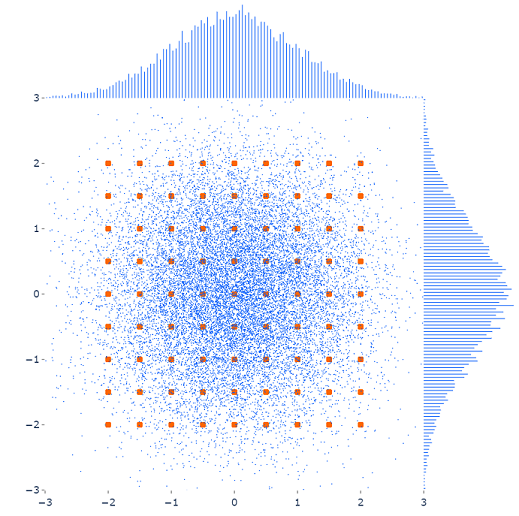}
\end{center}
    \captionsetup{belowskip=-10pt}
   \caption{Distribution of Shift Values (Axes in mm)}
\label{fig:shift_values}
\end{figure}

To address the limited realism of the shift distribution employed in previously considered work, random variables were utilized to generate the input displacements for the shift-simulating cropping process. Namely, two random samples were chosen from univariate Gaussian distributions with equivalent parameterizations. Parameters were chosen such that the resulting shift  values mimicked those anticipated to be generated through the proposed mechanism of manual HMD repositioning. The resulting distribution of shift values used herein is depicted in Fig. \ref{fig:shift_values}, along with the previously considered rectangular grid values.
To promote comparison with prior studies, initial simulations were conducted using Gaussian distributions with zero mean and 1 mm standard deviation. This latter parameter was selected to ensure that approximately 95\% of simulated shifts were contained within the previously considered $\pm$ 2 mm range. 

As discussed in Section 4.3, additional analysis was performed for shifts generated using a zero mean distribution with a standard deviation of 2.5 mm. This investigation was completed to assess performance of the proposed technique for realistic shift values extending beyond those previously considered in the literature. The standard deviation parameter was chosen based upon manual examination of the output images of the cropping process. Shifts extending beyond 5 mm were sometimes found to generate images with eye elements placed beyond the boundaries. As this represents a fundamental limitation of our simulation workflow, we have limited our analysis to this shift range. Since larger values may be encountered during use, further exploration is required to consider these scenarios. 
\subsection{Machine Learning}
\begin{figure*}[tb]
\begin{center}
   \includegraphics[width=.7\linewidth]{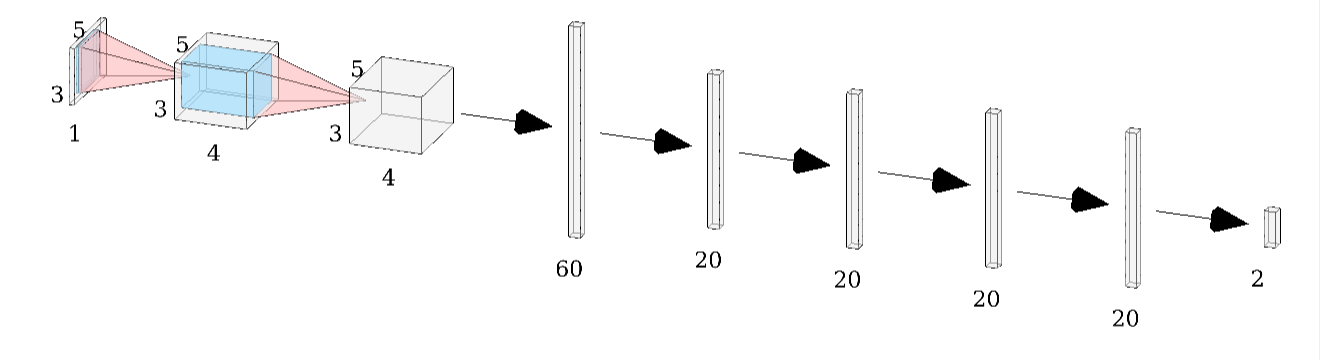}
\end{center}
   \caption{Low Power CNN Architecture}
\label{fig:LPCNN}
\end{figure*}

Of the various mapping architectures considered in~\cite{katrychuk_power-efficient_2019}, only the low-power CNN model was analyzed herein. This focus is motivated by the desired application within wireless HMDs, which informed the choice of constraints used to select network parameters in the grid search procedure. The CNN consists of two convolutional layers (each with output channel size of four), followed by four fully connected layers (20 neurons per layer) as shown in Fig. \ref{fig:LPCNN}.

Both the aforementioned transfer learning (hereby referred to as fine-tuning (FT)) and individual-specific (hereby referred to as from scratch (FS)) training approaches were simulated within this analysis. The FS training approach uses data collected from the recording session of a unique individual for training, with initial network weights set randomly. The FT approach initializes network weights through pre-training on the entire out-of-subject data pool. This procedure is slightly modified from that described in ~\cite{katrychuk_power-efficient_2019}, where pre-training was conducted on only a batch of out-of-subject data. The leave-one-subject-out pre-training strategy employed herein was chosen to maximize the use of available training data for the initialization of network weights.  

\subsection{Experimental Structure}
A dedicated experiment was conducted to assess variability in network performance as a function of the amount of training data utilized. The entire subject-specific dataset was initially partitioned into a 60\%/10\%/30\% train/validation/test split, with the initial partition hereby referred to as the training superset. Mapping networks were then trained using both strategies (i.e.: FS and FT) for partitions of this training superset, ranging from 20\% to 100\% in increments of 20\%. As referenced in Section 4.1, this partitioning strategy allows for comparison with the prior results presented in ~\cite{katrychuk_power-efficient_2019}, where a 24\%/6\%/70\% train/validation/test split was used. 

An additional experiment was conducted to assess the ability of the network to accommodate shift ranges outside of those encountered in training. This was accomplished by first performing a 56\%/14\%/30\% train/validation/test split for all samples of shifts less than 1.0 mm. Shifts of magnitude exceeding 1.0 mm were placed into dedicated testing bins as a function of shift range. This procedure produced four dedicated test sets containing the following exclusive shift ranges – 1) [0.0, 1.0], 2) (1.0, 1.5], 3) (1.5, 2.0], and 4) $>$ 2.0 mm. Elements in bin 1 represent shift magnitudes encountered in the training set, while bins 2 - 4 correspond to shifts outside of the those included in training. 

A final experiment was conducted to assess network performance for an expanded range of shift values. Shift inputs were generated using the aforementioned Gaussian distributions with  standard deviations of 2.5 mm. Data generated using these distributions was segmented using the same train/validation/test split (i.e.: 24\%6\%10\%) as used in ~\cite{katrychuk_power-efficient_2019}. 

\section{Results and Discussion}
\subsection{Performance Variability Versus Scale of \\Available Training Data}
\begin{figure}[t]
\begin{center}
   \includegraphics[width=\linewidth]{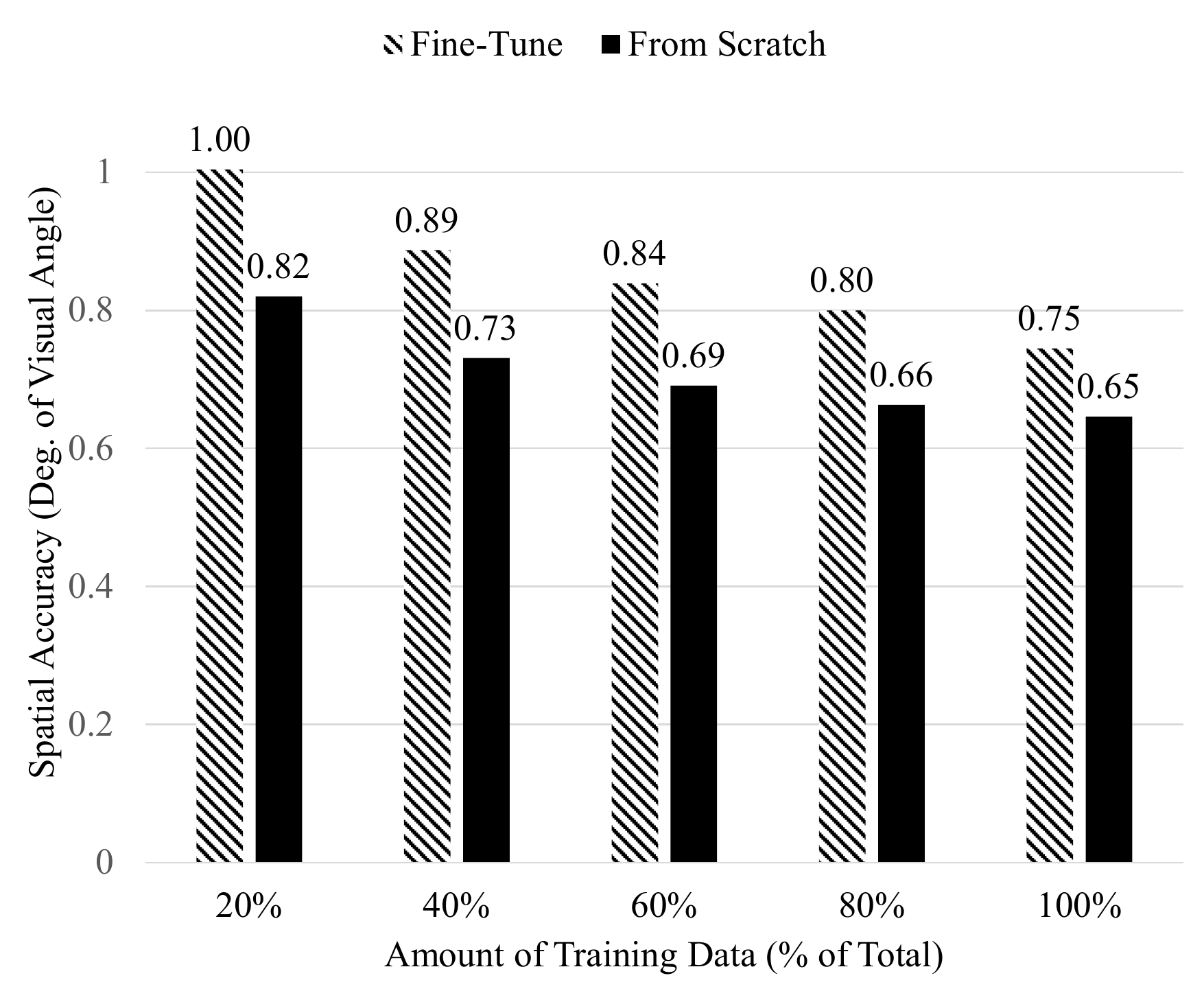}
\end{center}
   \caption{Mean Spatial Accuracy Versus Available Training Data}
\label{fig:data_scale}
\end{figure}

Variation in mean spatial accuracy across subjects versus available training data is depicted in Fig. \ref{fig:data_scale}. For both training methods considered, spatial accuracy is monotonically decreasing (i.e.: increasing system performance according to the definition of this metric), with network generalization improving with further exposure to training data as expected. For purposes of comparison, the mapping network produced a spatial accuracy of 1.07$^{\circ}$ (FT) and 0.77$^{\circ}$ (FS) with 4,273 average training examples using the less realistic uniform rectangular grid to simulate shift values in  ~\cite{katrychuk_power-efficient_2019}. For the more-realistic distribution of spatial distributions introduced herein, a performance of 0.89$^{\circ}$ (FT) and 0.73$^{\circ}$ (FS) was achieved for an equivalent amount of training data (note that 40\% of the 60\% training superset yields an equivalent training percentage of 24\% as was used in ~\cite{katrychuk_power-efficient_2019}). 

For the smallest subset of training data considered, results are comparable to previously demonstrated solutions (i.e.: PS-V). These observations may be exploited to reduce set-up duration for applications in which marginal accuracy degradations are acceptable. For example, under the assumption of a constant training data collection rate, reducing the initialization duration by 80\% results in a relative accuracy reduction of only 34.8\% (FT) and 26.9\% (FS).

\begin{figure*}[t]
\begin{center}
   \includegraphics[scale=0.8]{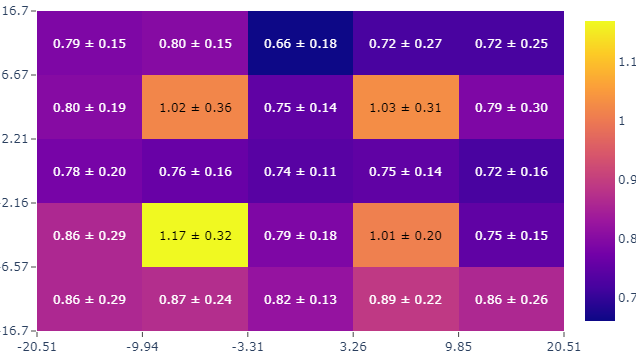}
\end{center}
   \caption{Spatial Accuracy Map (Boundary Values Represent Gaze Location, Interior Values Represent Mean Spatial Accuracy $\pm$ One Standard Deviation. Both Quantities are Measured in Degrees of the Visual Angle}
\label{fig:spat_acc_map}
\end{figure*}

Variation in spatial accuracy (i.e.: mean $\pm$ one standard deviation) across the operating range of the device is shown in Fig. \ref{fig:spat_acc_map}. These values were generated using a 40\% partition of the training superset, with training accomplished using the FS method. Reported values are obtained by discretizing the test set in the spatial domain using the binning structure introduced in Fig. \ref{fig:DataDist} for each subject, and then averaging across subjects. Spatial accuracy is worst for the four diagonal bins located off of center. This may be explained through examination of the stimulus and gaze locations depicted in Fig. \ref{fig:DataDist}. Namely, as no fixations were located within these bins (i.e.: gaze locations that likely correspond to saccadic eye movements), the scale of available testing and training data is minimal versus those bins where stimulus fixations were located. 

\subsection{Performance Variability For Out-of-Training Shifts}
\begin{figure}[t]
\begin{center}
   \includegraphics[width=\linewidth]{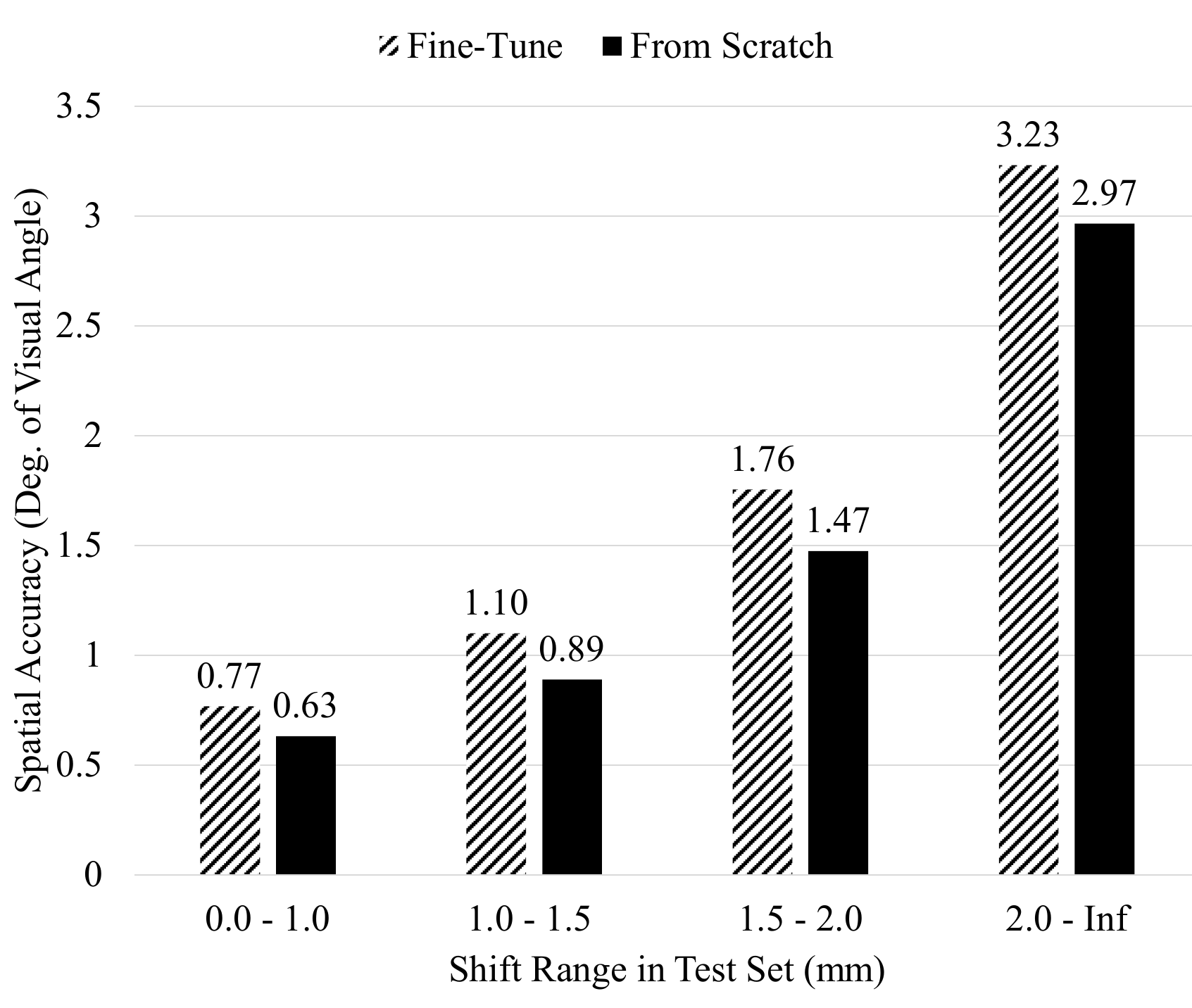}
\end{center}
    \captionsetup{belowskip=-10pt}
   \caption{Mean Spatial Accuracy Versus Shift Range}
\label{fig:shifts_range}
\end{figure}

Network performance for both in- and out-of-training shifts is depicted in Fig. \ref{fig:shifts_range}. For both training methods considered, spatial accuracy degrades when out-of-training shifts are introduced in testing as expected. Shifts of magnitude slightly exceeding those encountered in training produce some accuracy degradation. For the (1.0, 1.5] test set, which corresponds to a maximum shift extension of 0.5 mm beyond the training set, spatial accuracy is reduced by 0.33$^{\circ}$ (FT) and 0.26$^{\circ}$ (FS) versus the in-training benchmark.

As the range of test shifts increases, performance degradation is more considerable. For shifts exceeding 2.0 mm (with an upper bound dictated by the roll-off of the random distributions used as input to the shift simulating process), performance degrades by 2.46$^{\circ}$ (FT) and 2.34$^{\circ}$ (FS) against the in-training benchmark. The maintenance of reasonable spatial accuracy for slight out-of-training shifts is promising, as such scenarios may arise in practical deployment, especially when the initialization duration is limited to enhance user convenience.

\subsection{Performance for Extended Shift Range}
For the experiment exploring an expanded range of sensor shifts, a spatial accuracy of 1.31 $^{\circ}$ (FT) and 1.18 $^{\circ}$ (FS) was achieved using the two training methods considered. The improved accuracy using the FS approach is consistent with the other experiments conducted herein. An exploration of the potential source of performance discrepancies across training strategies is provided in the following section. 

These results indicate that reasonable performance may be maintained for shifts larger than those previously considered within the literature, assuming the network is exposed to such values during the training process. As noted in Section 3.2, limitations of the image-based workflow employed herein do not allow for the assessment of shifts extending beyond 5 mm.

\subsection{Comparative Performance for Different \\Training Approaches}
As demonstrated in the prior subsections, spatial accuracy for FS training is superior to the FT approach for each simulation considered. The source of this discrepancy is not immediately obvious, as both the FS and FT techniques utilize an identical subject-specific dataset within the training process, differing only in the technique utilized to initialize network weights. Additional improvements to the FT approach are of particular value for the target application, given its previously demonstrated ability to accelerate training convergence ~\cite{katrychuk_power-efficient_2019}. 

\begin{figure}[t]
\begin{center}
   \includegraphics[width=\linewidth]{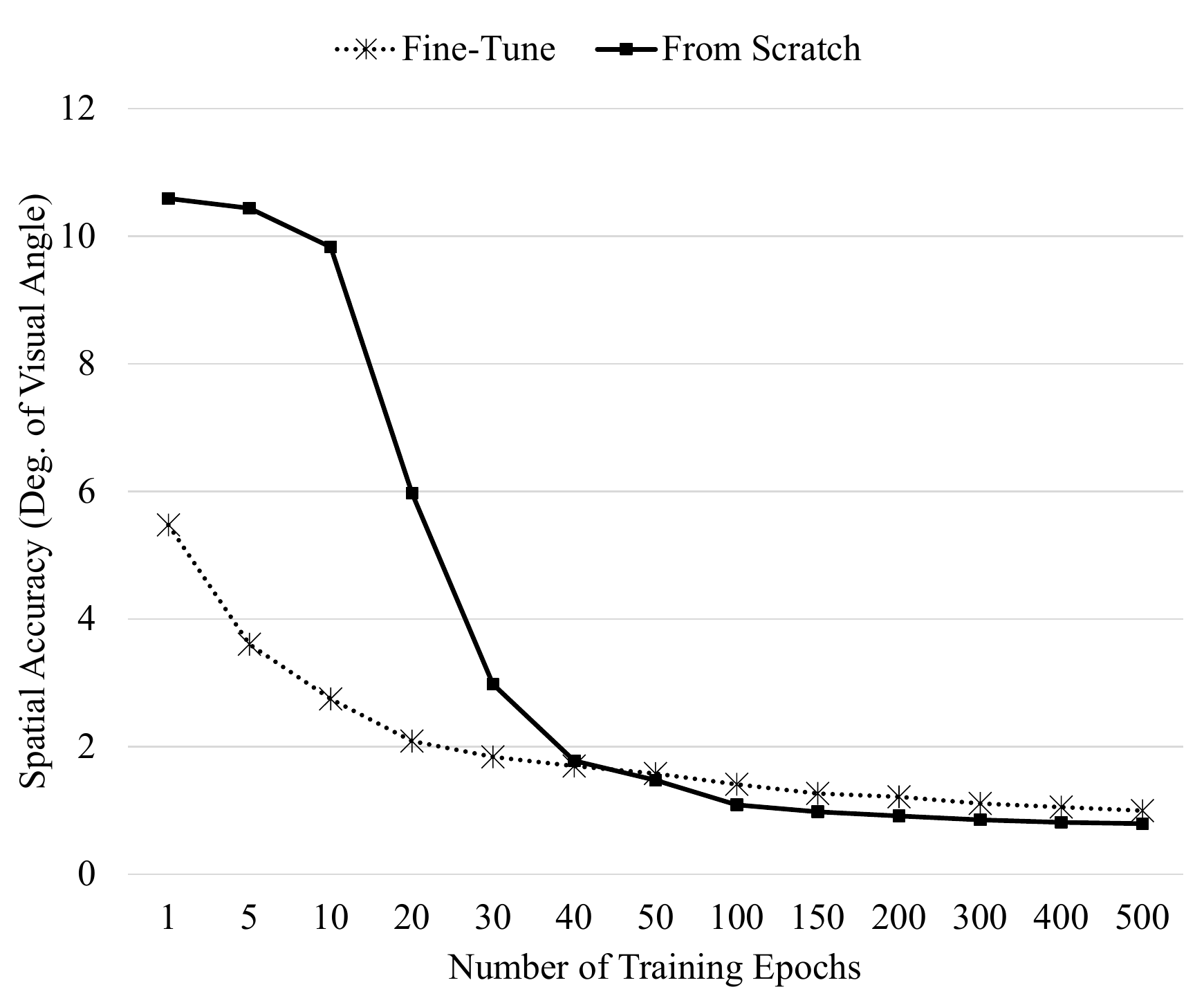}
\end{center}
    \captionsetup{belowskip=-10pt}
   \caption{Mean Spatial Accuracy Versus Number of Training Epochs}
\label{fig:ft_fs_epochs}
\end{figure}

To further assess the training efficiency of the FT technique, spatial accuracy was computed for a varying number of training epochs for both the FS and FT method. As demonstrated in Fig. \ref{fig:ft_fs_epochs}, the FT technique produced superior accuracy versus FS through the first 50 training epochs due to the transfer of knowledge from the pre-training pool. At approximately 50 epochs, FS training begins to exhibit superior performance.

\begin{figure*}[tb]
\begin{center}
   \includegraphics[width=.85\linewidth]{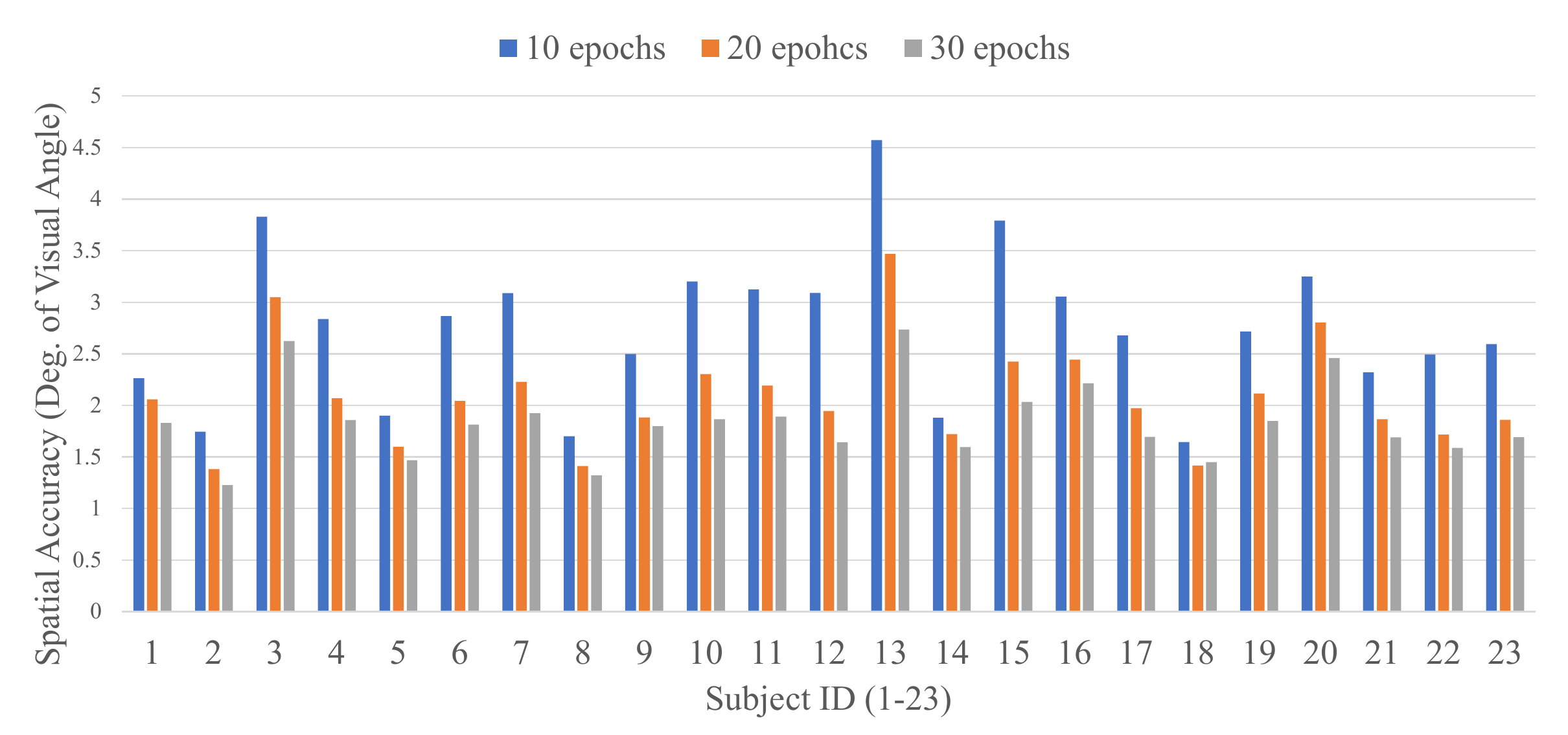}
\end{center}
    \captionsetup{aboveskip=-10pt}
   \caption{Variation in Spatial Accuracy Across Subjects for Multiple Training Duration} 
\label{fig:SSEpoch}
\end{figure*}

Spatial accuracy for the FT technique is presented in Fig. \ref{fig:SSEpoch} on a per-subject basis for a varying number of training epochs at the initialization of user-specific training. 
As noted, variability amongst subjects is considerable (i.e.: ranging from 1.64$^{\circ}$ to 4.57$^{\circ}$), indicating that the accuracy afforded by pre-training is highly person-specific (i.e.: the level of generalization of gaze maps learned out-of-subject varies across individuals). Additionally, the rate of accuracy improvement upon exposure to subject-specific training data also exhibits significant variability across the subject pool.  

Future work will attempt to reduce the performance discrepancy between the two training techniques. One possible approach of interest is the restriction of the FT pre-training pool to a subset of individuals exhibiting anatomical similarity to the user. It is hypothesized that this technique may reduce the inter-subject variability in the initial accuracy achieved by the FT approach during training. The ultimate goal of this effort is to produce a reasonable level of accuracy through pre-training alone, thereby allowing for the possibility of off-the-shelf use without user-specific customization.

\subsection{Limitations}
While the presented results are promising for advancing the understanding of the gaze mapping network, they are characterized by several limitations. Simulated shifts using the current processing workflow are restricted to translations in two-dimensions only, whereas practical shifts will involve more complex displacements (i.e.: slippage along the nose, rotations, etc.). Furthermore, the observed signal quality in hardware will be degraded versus the simulations performed herein using a high-quality image. Additionally, a full assessment of the trade-off between accuracy and initialization duration must also analyze the spatial distribution, display duration, and number of targets used during training. Finally, the feasibility of the proposed manual displacement technique for inducing shift training data must be thoroughly tested using prototype hardware.

\section{Conclusions}
The research described herein demonstrates the ability of a CNN to provide shift-invariant gaze mapping for PS-OG EM sensors in the presence of more realistic shift training data. Simulated shifts were generated using coordinate values sampled from two normal distributions. This method was chosen to better reflect the spatial variability of shifts which would be realized in practice through manual translations of an HMD by the end-user during a calibration procedure. Observed spatial accuracy was comparable to prior results obtained using an unrealistic rectangular shift grid. This result suggests that the proposed mechanism of generating training data through manual HMD repositioning may be feasible in practical scenarios.

Variability in network performance versus training data scale was also assessed. Results indicate that reasonable accuracy may be achieved with limited training data,  allowing for an operational trade-off in which initialization duration is reduced for applications with less stringent accuracy requirements. The ability of the network to generalize to shifts extending beyond those observed during training was also demonstrated. While achieved accuracy degrades considerably for testing samples with significantly out-of-range shifts, best-case spatial accuracy was maintained at less than one degree of the visual angle for shift ranges extending up to 50\% beyond those encountered in training.

\section*{Acknowledgements}
This work is supported by the National Science Foundation under Grant Numbers CNS-1250718 and CNS-1714623. The work is inspired by the Google Virtual Reality Research Award and Google Global Faculty Research Award bestowed on  Dr. Komogortsev in 2017 and 2019. 

{\small
\bibliographystyle{ieee_fullname}

}

\end{document}